\documentclass[11pt]{llncs}        
 
\usepackage{fullpage}    
\usepackage{proof}
\usepackage{times}
\usepackage{rotating}
\usepackage{amssymb}
\usepackage{amstext}
\usepackage{pst-node,pst-tree}

\newpsobject{showgrid}{psgrid}{subgriddiv=1,griddots=0,gridlabels=6pt}

\newcommand{\myparagraph}[1]{\vspace*{2mm}\noindent \textbf{#1}}
                                  
\newcommand{\Vampire}{Vampire}
\newcommand{\E}{E}
\newcommand{\Waldmeister}{Waldmeister}
\newcommand{\SPASS}{Spass}
\newcommand{\Gandalf}{Gandalf}
\newcommand{\OTTER}{Otter}
\newcommand{\DISCOUNT}{DISCOUNT}
\newcommand{\SCOTT}{SCOTT}
\newcommand{\SETHEO}{SETHEO}
\newcommand{\Octopus}{Octopus}
\newcommand{\Theo}{Theo}
\newcommand{\SNARK}{SNARK}
\newcommand{\inc}{~~\= \+ \kill}
\newcommand{\dinc}{\inc\inc}
\newcommand{\dec}{\- \kill}
\newcommand{\ddec}{\dec\dec}
\newcommand{\reserved}[1]{\textbf{\underline{#1}}}

\newcommand{\Active}{\mathit{active}}
\newcommand{\Passive}{\mathit{passive}}
\newcommand{\Current}{\mathit{current}}

\newcommand{\Input}{\mathit{input}}
\newcommand{\New}{\mathit{new}}

\newcommand{\Select}{\mathit{select}}
\newcommand{\Infer}{\mathit{infer }}

\begin{document}

\mainmatter

\title{New Implementation Framework for Saturation-Based Reasoning}

\author{Alexandre Riazanov}

\institute{
    \email{alexandre.riazanov@gmail.com}\\ \vspace{3mm}
    December 12, 2006
}

\maketitle

\begin{abstract}
The \emph{saturation-based reasoning methods} are among the most theoretically
developed ones and are used by most of the state-of-the-art
first-order logic reasoners. In the last decade there was 
a sharp increase in performance of such systems, which I attribute
to the use of advanced calculi and the intensified research in 
implementation techniques. However, nowadays we are witnessing
a \emph{slowdown in performance progress}, which may be considered as 
a sign that the saturation-based technology is reaching 
its inherent limits. The position I am trying to put forward in this
paper is that \emph{such scepticism is premature} and a sharp
improvement in performance may potentially be reached by adopting 
\emph{new architectural principles for saturation}.
The top-level algorithms and corresponding designs 
used in the state-of-the-art saturation-based
theorem provers have (at least) two inherent drawbacks:
the \emph{insufficient flexibility of the used inference selection 
mechanisms} and the \emph{lack of means for intelligent prioritising
of search directions}. In this position paper I analyse these drawbacks
and present two ideas on how they could be overcome.
In particular, I propose a 
\emph{flexible low-cost high-precision mechanism for inference selection}, 
intended to overcome problems associated with the currently used instances 
of clause selection-based procedures. I also outline a method for intelligent 
prioritising of search directions, based on \emph{probing the search space
by exploring generalised search directions}. I discuss some 
technical issues related to implementation of the proposed architectural
principles and outline possible solutions. 
\end{abstract}


                 \section{Introduction}


An automatic \emph{theorem prover} for first-order logic (FOL)
is a software system that can be used to show that some conjectures 
formulated in the language of FOL are implied by some theory. 
The expressiveness of FOL and its relative mechanisability 
make automated theorem proving
in FOL a useful instrument for such applications as verification 
\cite{Crocker:IJCAR:2001,ClaessenHahnleMartennsson:PaPS:2002,Ahrendt+:JELIA:2000,Detlefs+:COMPAQ:1998}
and synthesis \cite{Lowry+:NASA:1994}
of hardware and software,
knowledge representation \cite{Lenat:CyC:CommACM:1995}, 
Semantic Web \cite{HorrocksPatelSchneider:WWW:2003}, 
assisting human mathematicians 
\cite{McCune:JAR:RobbinsProblem:1997,Benzmueller+:Omega:CADE:1997}, 
background reasoning in interactive theorem provers
\cite{Meng:IJCAR:SupportingInteractiveProof:2004}, 
and others.

This paper is concerned with the theorem proving method 
based on the concept of \emph{saturation}.
Given an input set of formulas, the prover 
tries to \emph{saturate} it under all inferences in the inference system
of the prover. In order to deal with syntactic objects which 
allow efficient calculi,
the input set of formulas is usually converted into a set of formulas of
a special form, called \emph{clauses}
\footnote{Universally quantified disjuncts of literals. A literal is
either an atomic formula (possibly depending on some variables), or 
a negation of such an atomic formula.}. Demonstrating validity 
of a first-order formula is thereby reduced to demonstrating
unsatisfiability of the corresponding set of clauses
\footnote{
Sometimes problems coming from applications are already represented
in the clausal form or require only minor transformation.}. 
The calculi working with clauses are usually designed in such a way 
that inferences can only produce clauses 
(see, e.~g., \cite{BachmairGanzinger:HandbookAR:resolution:2001,Nieuwenhuis:HandbookAR:paramodulation:2001}).

There are three possible
outcomes of the saturation process on clauses: (1) an empty clause is derived,
which means that the input set of clauses is unsatisfiable; 
(2) saturation terminates without producing an empty clause, 
in which case the input set of clauses is satisfiable
(provided that a complete inference system is used); 
(3) the prover runs out of resources.
The saturation method is well-studied theoretically
(\cite{BachmairGanzinger:HandbookAR:resolution:2001,Nieuwenhuis:HandbookAR:paramodulation:2001}) 
and is implemented in a significant number of modern provers,
e.~g., \E\ \cite{Schulz:AICOM:brainiac:2002}, \E-\SETHEO\ (the \E\
component), \Gandalf\ \cite{Tammet:JAR:Gandalf:1997}, 
\OTTER\ \cite{McCune:Otter3.0:1994}, 
\SNARK\ \cite{Stickel:SNARK:2005},
\SPASS\ \cite{Weidenbach:HandbookAR:SPASS:1999}, 
\Vampire\ \cite{RiazanovVoronkov:design_and_implementation_of_Vampire:AICOM:2002,Riazanov:PhDThesis:2003},
and \Waldmeister\ \cite{Hillenbrand+:AICOM:phytography:2002}.

In the last decade there has been a \emph{sharp increase in performance}
of such systems\footnote{
A good benchmark is \OTTER\, which has not changed much
since 1996. 
Compare its relative performance in CASC-13 
(http://www.cs.miami.edu/$\sim$tptp/CASC/13/) and 
CASC-20 (http://www.cs.miami.edu/$\sim$tptp/CASC/20/). 
}, 
which I attribute to the use of advanced calculi and inference
systems (primarily, complete variants of resolution
\cite{BachmairGanzinger:HandbookAR:resolution:2001}
and paramodulation
\cite{Nieuwenhuis:HandbookAR:paramodulation:2001} 
with ordering restrictions, and a number of compatible redundancy 
detection and simplification techniques),
and intensified research on efficient implementation techniques, 
such as term indexing (see \cite{Graf:TermIndexing:1996} and 
more recent survey \cite{SekarRamakrishnanVoronkov:HandbookAR:Indexing:2000}), 
heuristic methods for guiding proof search (see, e.~g., 
\cite{Schulz:AICOM:brainiac:2002})
and top-level saturation algorithms (see, e.~g., 
\cite{HL02} and \cite{RiazanovVoronkov:JSC:LRS:2002}). 
Unfortunately, the initial momentum created by such work
seems to have diminished, 
and nowadays we are witnessing
a slowdown in performance progress\footnote{
 Compare the performance of the best provers in 
CASC-20 (http://www.cs.miami.edu/$\sim$tptp/CASC/20/)
with the previous year winners. 
}. 
Some researchers consider this to be a sign that
the saturation-based reasoning technology is reaching 
its inherent limits. 
The position I am trying to defend in this paper is
that \emph{such scepticism is premature}. My argumentation is based
on a thesis that potential opportunities for a new breakthrough 
in performance have not been exhausted. Namely, the possibility
of adopting \emph{new implementation frameworks} for saturation, 
i.e., top-level designs and algorithms, has not been fully explored.
To support this claim, I will pinpoint some major weaknesses in 
the organisation of proof search in the standard approaches to implementing 
saturation, and propose two concrete ideas on how to overcome these
problems.

First, I will analyse some inherent problems with the standard procedures
for saturation, based on the implementation of \emph{inference selection 
via clause selection}. In particular, I consider the two main
procedures, the OTTER algorithm and the DISCOUNT algorithm,
\emph{based on clause selection}. The main problem with the former
procedure is the \emph{coarseness of inference selection}, which translates
into \emph{insufficient productivity of heuristics and restricts the choice
of possible heuristics}. The latter procedure implements very fine
selection of inferences, but at a \emph{high cost} in terms of computational
resources. I will propose
a new procedure based on a \emph{flexible high-precision inference selection}
mechanism with \emph{acceptable overhead}. A concrete implementation scheme
will be outlined. 

Second, I will highlight the inadequacy of the popular approaches
to prioritising proof search directions, based on \emph{syntactic 
characteristics} of \emph{separate} clauses. As a possible remedy,
I propose a method for intelligent prioritising of search directions,
based on \emph{probing the search space by exploring generalised 
search directions}. I also propose a concrete implementation scheme
for the method.

This criticism of the current state of affairs in the saturation
architectures originates in my hands-on experience 
with implementing the saturation-based kernel of 
\Vampire\ \cite{RiazanovVoronkov:design_and_implementation_of_Vampire:AICOM:2002,Riazanov:PhDThesis:2003}, 
and numerous experiments with the system. In fact, I consider the observations
related to proof search effectiveness, on which this paper is based,
the most valuable lessons learned from the Vampire kernel implementation.
However, this paper is only 
a position paper. As such, it does not present any complete results, either
theoretical or experimental. Its aim is to provide a basis and an inspiration
for new implementations and experiments.

The rest of this paper is structured as follows. Each of the 
remaining two sections introduces a new architectural principle. 
In the beginning of the chapter the relevant aspects of the
state-of-the-art designs are criticised. Then, ideas 
of a possible remedy are formulated, followed by
a discussion of related work and a tentative research programme. 

Concluding this introduction, I would like to ask the reader to be tolerant to
some presentational problems with this text. 
I am trying to keep this paper informative
for experts in the implementation of saturation-based provers 
and, at the same time, acceptable for a superficial reading 
by a broader audience. Some negative consequences of such 
conflict of intentions seem to be inevitable.


                 \section{General preliminaries}


For the sake of self-containedness, I will reproduce a number of standard 
definitions here.

I am assuming that the reader is familiar with the syntax and semantics 
of first-order predicate logic with equality. In what follows, ordinary
predicate symbols will be denoted by $p$, $q$ and $r$, the equality
predicate will be denoted by $\simeq$, 
function symbols will be denoted by $f$, $g$ and $h$,
individual constants will be denoted by $a$, $b$ and $c$, 
variables will be denoted by $x$, $y$ and $z$, possibly with subscripts,
and the letters $s$ and $t$, possibly with subscripts, will denote terms.

We are mostly interested in a special kind of first-order formulas called
\emph{clauses}. A clause is a disjunction $L_1 \vee \ldots L_n$, where 
all $L_i$ are \emph{literals}, i~.e. atoms (positive literals) or
negated atoms (negative literals). The order of the literals in a clause
is usually irrelevant, so I will often refer to clauses as finite 
multisets of literals. The empty multiset of literals will also be 
considered a clause which is false in any interpretation.

\emph{Substitutions} are total functions that map variables to terms.
They will be denoted by $\theta$ and $\sigma$, possibly with subscripts.
Substitution application is extended to complex expressions, such as terms,
atoms, literals and clauses, in an obvious way: if $E$ is an expression,
$E\theta$ is obtained by replacing each variable $x$ in $E$ by $x\theta$. 
A substitution $\theta$ is a \emph{unifier} for two expressions $E_1$ and
$E_2$ if $E_1\theta = E_2\theta$. It is the \emph{most general unifier}, 
if for any other unifier $\theta_1$, there exists a substitution $\theta_2$, 
such that $E_1\theta_1 = (E_1\theta)\theta_2$.

We will say that a clause $C$ \emph{subsumes} clause $D$ if there is a 
substitution $\theta$ such that (the multiset of literals) $C\theta$ 
is a submultiset of $D$. 

We are interested in implementation of calculi based on \emph{resolution}
and \emph{paramodulation} (see, e.~g., \cite{BachmairGanzinger:HandbookAR:resolution:2001,Nieuwenhuis:HandbookAR:paramodulation:2001}).
(Unrestricted) \emph{binary resolution} is the following deduction rule:
\begin{center}    	
 \[
      \infer
       {(C \vee D)\theta}
       {C \vee A~~~D \vee \neg B}
    \]
 where $\theta$ is the most general unifier of the atoms $A$ and $B$.
\end{center} 
Paramodulation is the following rule:
\begin{center}    	
 \[
      \infer
       {(C \vee D[t])\theta}
       {C \vee s \simeq t~~~D[u]}
    \]
 where $\theta$ is the most general unifier of the terms $s$ and $u$, and
 $u$ is not a variable.
\end{center} 
A resolution-based reasoner usually applies restricted variants of these rules 
together with some auxiliary rules to demonstrate unsatisfiability of
an input clause set by deriving an empty clause from it.
Such derivations are called \emph{refutations} of the corresponding 
clause sets. 

Saturation-based reasoners are called so because of the way they search
for refutations. In an attempt to derive an empty clause, a reasoner
tries to \emph{saturate} the initial set with all clauses derivable from 
it. Roughly speaking, at some steps of the \emph{saturation process} 
the reasoner \emph{selects a possible inference} between some clauses
in the current clause set, applies the inference and adds the resulting
clause to the current clause set. Other steps of the process usually
prune the search space by removing \emph{redundant} clauses, i.~e. clauses
that are not strictly necessary to find a refutation. For details on the
concept of saturation modulo redundancy, the reader is referred to 
\cite{BachmairGanzinger:HandbookAR:resolution:2001}.


                 \section{Fine inference selection at affordable cost}


\subsection{Background: inference selection via clause selection}
      \label{subsec:inference_selection_via_clause_sel}

When one has to search in an indefinitely large space, 
the ability to \emph{explore more promising search directions
before the less promising ones} is a key to success.
In saturation-based reasoning the mechanism responsible for
deciding on which direction to promote first is
known as \emph{inference selection}. Ideally,
the inference selection should be able to name 
\emph{one single inference} to be deployed at every step 
of saturation, and the decision should be based on the 
(heuristically evaluated) quality
of the resulting clause. In practice, most of the working 
saturation-based systems adopt a simpler but coarser mechanism 
known as \emph{clause selection}.
Instead of selecting a single inference at a time, 
we \emph{select a clause} and \emph{oblige} to deploy immediately \emph{all} possible
inferences between the clause and \emph{all active} (previously 
selected) clauses. Clauses of better heuristically evaluated quality
are given higher priority for selection, in the hope that they
will produce heuristically good inferences.
The algorithm realising inference selection via
clause selection is known as \emph{given-clause algorithm}. 
Its variants have been used in provers \emph{since as early as 1974} 
\cite{Overbeek:JACM:ATPalgorithms:1974} 
(see also \cite{Lusk:LPAR:ArgonneStyle:1992}), 
although its current 
monopoly seems to be mostly due to the success of \OTTER\ 
\cite{McCune:Otter3.0:1994}. Other provers based on variants of given-clause 
algorithm include \E, \Gandalf, \SNARK, \SPASS, \Vampire\ and \Waldmeister,
i.~e., practically all modern saturation-based systems. 

In order to illustrate the main idea behind 
the given-clause algorithm, namely the implementation of inference selection
via clause selection, it is sufficient to consider only deduction inferences.
So, the algorithm presented in Figure~\ref{fig:GivenClauseAlgorithm} performs
no simplification steps.


\begin{figure}
\begin{center}
\begin{minipage}[t]{150mm}
\begin{tabbing}
 \reserved{procedure} $\mathit{GivenClause}$($\Input$ : set of clauses) \\  \dinc
 \reserved{var} $\New$, $\Passive$, $\Active$ : sets of clauses  \\
 \reserved{var} $\Current$ : clause  \\
 $\Active\ := \emptyset$  \\
 $\Passive\ := \Input$ \\
 \reserved{while} $\Passive \neq \emptyset$ \reserved{do} \\ \dinc
  $\Current\ := \Select(\Passive)$  \\
  $\Passive\ := \Passive - \{\Current\}$  \\
  $\Active\ := \Active \cup \{\Current\}$  \\
  $\New\ := \Infer(\Current,\Active)$  \\
  \reserved{if} $\New$ contains empty clause \\ \dinc
    \reserved{then} \reserved{return} \textit{refutable}  \\ \ddec
  $\Passive\ := \Passive \cup \New$  \\ \ddec
\reserved{od}  \\
\reserved{return} \textit{failure to refute}
\end{tabbing}
\end{minipage}
                   \caption{Given-clause algorithm (without simplifications)
                         \label{fig:GivenClauseAlgorithm}}
\end{center}
\end{figure}



\begin{figure*}
\begin{center}
\psset{xunit=.09cm,yunit=.05cm,dash=2pt 2pt}
\begin{pspicture}(0,-75)(150,0)

\psframe[framearc=.6](10,-5)(50,-20)
\rput(30,-12.5){$\mathit{input~clauses}$}
\pnode(30,-20){input_out}

   \psframe[framearc=1.0](60,-9)(100,-16)
   \rput(80,-12.5){$\Current$}
   \pnode(65,-16){current_in}
   \pnode(80,-16){current_out1}
   \pnode(100,-12.5){current_out2}

\psframe[framearc=.6](10,-35)(50,-50)
\rput(30,-42.5){$\Passive$}
\pnode(30,-35){passive_in1}
\pnode(10,-42.5){passive_in2}
\pnode(30,-50){passive_in3}
\pnode(50,-40){passive_out1}
\pnode(50,-45){passive_out2}

   \psframe[framearc=.6](60,-35)(100,-50)
   \rput(80,-42.5){$\Active$}
   \pnode(80,-35){active_in1}
   \pnode(60,-40){active_in2}
   \pnode(100,-40){active_out1}
   \pnode(80,-50){active_out2}

       \psframe(110,-35)(150,-50)
       \rput(130,-42.5){$\mathit{deduction~inf.}$}
       \pnode(130,-50){infer_out}
       \pnode(130,-35){infer_in1}
       \pnode(110,-40){infer_in2}

     \psframe[framearc=1.0](110,-69)(150,-76)
     \rput(130,-72.5){$\New$}
     \pnode(130,-69){new_in1}
     \pnode(130,-76){new_in2}
     \pnode(110,-72.5){new_in3}
     \pnode(110,-72.5){new_out}

\pnode(30,-72.5){aux_node1}

\ncline{->}{input_out}{passive_in1}\ncput{~~~1}
\ncline{->}{passive_out1}{current_in}\ncput{~~~~2}
\ncline{->}{current_out1}{active_in1}\ncput{~~~~3}
\ncline{->}{infer_out}{new_in1}\ncput{~~~~4}
\ncline{-}{new_out}{aux_node1}
\ncline{->}{aux_node1}{passive_in3}\ncput{~~~~5}

\ncline[linestyle=dashed]{->}{current_out2}{infer_in1}\ncput{~~~~~~4}
\ncline[linestyle=dashed]{->}{active_out1}{infer_in2}\ncput{\parbox[c][20pt][t]{5pt}{4}}

\end{pspicture}
                   \caption{Dataflow in the given-clause algorithm
                         \label{fig:GivenClauseAlgorithmDataflow}}

\end{center}
\end{figure*}
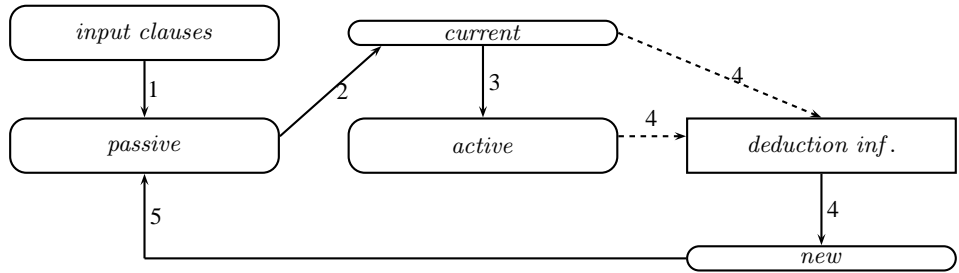

It is also convenient to represent the algorithm with a more abstract dataflow
diagram as in Figure~\ref{fig:GivenClauseAlgorithmDataflow}. In this picture, 
the boxes denote operations performed on clauses. The rounded boxes denote 
sets of clauses. The shallow ones correspond to the sets that typically contain
very few clauses, while the deep ones correspond to the sets that can grow large.
The arrows reflect the information flow for different operations. Arrows labeled 
with the same number belong to the same operation/processing phase. 
In Figure~\ref{fig:GivenClauseAlgorithmDataflow}, label~1 corresponds to the line
$passive := input$ in the pseudocode from Figure~\ref{fig:GivenClauseAlgorithm},
phase~2 is clause selection ($current := select(passive)$ and 
$passive := passive - \{current\}$), 3~corresponds to 
$active := active \cup \{current\}$, 4~is the generation of deduction inferences
between $current$ and $active$ ($new := infer(current,active)$),
and 5~is the integration of newly derived clauses into 
$passive$ ($passive := passive \cup new$). The thin solid arrows show the
movements of clauses between clause sets and from operations to the sets.
A dashed arrow from a set to an operation indicates that the operation
depends on the clauses from the set.

My experience with \Vampire\ and, to some extent, with other provers 
allows me to see a number of soft spots of the given-clause algorithm:

\begin{itemize}

\item The selection is based on the \emph{properties of eligible clauses}, 
which are only \emph{vaguely related to the properties
of the enabled inferences}. 

A ``good'' clause may interact with many
previously selected ``not-so-good'' clauses and produce many ``not-so-good''
inferences. The set of selected clauses often contains such heuristically
bad clauses for a number of reasons. In particular, we cannot completely
avoid selecting bad clauses because in general it leads to incompleteness.
Moreover, in practice we often cannot even significantly restrict the selection
of heuristically bad clauses since such strategy easily leads to loss
of solutions (in the practical sense, i.e., solutions that can be obtained
with given resources).
Another reason why bad clauses get into the set
of selected clauses is the relativity of the heuristic estimation 
of clause quality: a clause selected as relatively good in 
the beginning of the proof search, can become relatively bad later 
if many better clauses have been derived.

Another problem with 
clause property-based selection is that 
even two ``good'' clauses can easily have ``not-so-good'' inferences 
between them. This happens when the clause quality criteria do not
sufficiently penalise clauses containing ``bad'' parts available for inferences.
If our quality criteria are too strict with respect to clauses with
``bad'' parts, the prover also postpones the inferences involving 
``good'' parts of such clauses.  

\item The newly selected clause may, and often does, interact
with \emph{very many} parts of \emph{very many} active clauses. This
often leads to pathological situations of the following kind:
a prolific clause is selected and the processing of inferences 
between this clause and many active ones 
takes all available time, whereas a few inferences with other
clauses would lead to a solution. 

\end{itemize}

In sum, the \emph{coarseness} of the clause selection principle
\emph{deprives us of control over the proof search process
to a great extent}, which translates into \emph{poor productivity
of heuristics}, \emph{restricts the choice of heuristics} 
that can be implemented, and leads to 
\emph{littering the search state with too many ``undesirable'' clauses}. 

There are two main variants of the given-clause algorithm:
the \OTTER\ algorithm\footnote{
Implemented, in particular, in \Gandalf, \OTTER, \SNARK, \SPASS\ and \Vampire.
}
and the \DISCOUNT\ algorithm\footnote{
Implemented, in particular, in \E, \Vampire\ and \Waldmeister.
}, which differ in the way
the \emph{passive} (waiting to be selected) clauses are treated.


\begin{figure*}
\begin{center}
\psset{xunit=.09cm,yunit=.05cm,dash=2pt 2pt}
\begin{pspicture}(0,-110)(150,0)

\psframe[framearc=.6](10,-5)(50,-20)
\rput(30,-12.5){$\mathit{input~clauses}$}
\pnode(30,-20){input_out}

   \psframe[framearc=1.0](60,-9)(100,-16)
   \rput(80,-12.5){$\Current$}
   \pnode(65,-16){current_in}
   \pnode(80,-16){current_out1}
   \pnode(100,-12.5){current_out2}

\psframe[framearc=.6](10,-35)(50,-50)
\rput(30,-42.5){$\Passive$}
\pnode(30,-35){passive_in1}
\pnode(10,-42.5){passive_in2}
\pnode(30,-50){passive_in3}
\pnode(50,-40){passive_out1}
\pnode(50,-45){passive_out2}

   \psframe[framearc=.6](60,-35)(100,-50)
   \rput(80,-42.5){$\Active$}
   \pnode(80,-35){active_in1}
   \pnode(60,-40){active_in2}
   \pnode(100,-40){active_out1}
   \pnode(80,-50){active_out2}

       \psframe(110,-35)(150,-50)
       \rput(130,-42.5){$\mathit{deduction~inf.}$}
       \pnode(130,-50){infer_out}
       \pnode(130,-35){infer_in1}
       \pnode(110,-40){infer_in2}

\psframe(10,-65)(50,-80)
\rput(30,-72.5){$\mathit{backward~simpl.}$}
\pnode(40,-80){backwards_in}
\pnode(30,-80){backwards_out1}
\pnode(40,-65){backwards_out2}
\pnode(30,-65){backwards_out3}

   \psframe(60,-65)(100,-80)
   \rput(80,-72.5){$\mathit{forward~simpl.}$}
   \pnode(70,-65){forwards_in1}
   \pnode(80,-65){forwards_in2}
   \pnode(100,-72.5){forwards_out1}

     \psframe[framearc=1.0](110,-69)(150,-76)
     \rput(130,-72.5){$\New$}
     \pnode(130,-69){new_in1}
     \pnode(130,-76){new_in2}
     \pnode(110,-72.5){new_in3}
     \pnode(120,-76){new_out}

   \psframe[framearc=1.0](60,-99)(100,-106)
   \rput(80,-102.5){$\mathit{retained}$}
   \pnode(100,-102.5){retained_in}
   \pnode(60,-102.5){retained_out1}
   \pnode(70,-99){retained_out2}

\pnode(1,-102.5){aux_node1}
\pnode(1,-42.5){aux_node2}
\pnode(30,-110){aux_node3}
\pnode(130,-110){aux_node4}

\ncline{->}{input_out}{passive_in1}\ncput{~~~1}
\ncline{->}{passive_out1}{current_in}\ncput{~~~~2}
\ncline{->}{current_out1}{active_in1}\ncput{~~~~3}
\ncline{->}{infer_out}{new_in1}\ncput{~~~~4}
\ncline{->}{new_out}{retained_in}\ncput{~~~~6}

\ncline{-}{retained_out1}{aux_node1}
\ncline{-}{aux_node1}{aux_node2}\ncput{~~~~8}
\ncline{->}{aux_node2}{passive_in2}
\ncline{-}{backwards_out1}{aux_node3}\ncput{~~~~7}
\ncline{-}{aux_node3}{aux_node4}
\ncline{->}{aux_node4}{new_in2}\ncput{~~~~7}

\ncline[linestyle=dashed]{->}{current_out2}{infer_in1}\ncput{~~~~~~4}
\ncline[linestyle=dashed]{->}{active_out1}{infer_in2}\ncput{\parbox[c][20pt][t]{5pt}{4}}
\ncline[linestyle=dashed]{->}{passive_out2}{forwards_in1}\ncput{~~~~~5}
\ncline[linestyle=dashed]{->}{active_out2}{forwards_in2}\ncput{~~~~5}
\ncline[linestyle=dashed]{->}{retained_out2}{backwards_in}\ncput{~~~~7}

\ncline[doubleline=true]{->}{backwards_out3}{passive_in3}\ncput{~~~~7}
\ncline[doubleline=true]{->}{backwards_out2}{active_in2}\ncput{~~~~~7}
\ncline[doubleline=true]{->}{forwards_out1}{new_in3}\ncput{\parbox[c][20pt][t]{5pt}{5}}

\end{pspicture}
                   \caption{\OTTER\ algorithm
                         \label{fig:OTTERAlgorithm}}

\end{center}
\end{figure*}

In the \OTTER\ algorithm, presented as a dataflow diagram in Figure~\ref{fig:OTTERAlgorithm},
the passive clauses are subject to simplification
by the newly derived clauses, can be discarded as redundant 
with the help of the newly derived clauses,
and themselves can be used to simplify/discard
the newly derived clauses. Newly derived clauses are subject to 
\emph{forward simplification} which may transform them or even discard them 
completely \footnote{
In diagrams on Figures~\ref{fig:OTTERAlgorithm} and \ref{fig:DISCOUNTAlgorithm},
a broad arrow from an operation to a set indicates that the operation modifies
the set by removing or replacing some clauses.
} Note that in the \OTTER\ algorithm forward simplification uses both passive
and active clauses as simplifiers (see the dashed arrows labeled with 5 in 
the diagram). \emph{Backward simplification} also affects passive clauses
as well as the active ones (see the broad arrows labeled with~7).


\begin{figure*}
\begin{center}
\psset{xunit=.09cm,yunit=.05cm,dash=2pt 2pt}
\begin{pspicture}(0,-110)(150,0)

\psframe[framearc=.6](10,-5)(50,-20)
\rput(30,-12.5){$\mathit{input~clauses}$}
\pnode(30,-20){input_out}

   \psframe(60,-5)(100,-20)
   \rput(80,-10){$\mathit{backward~simpl.}$}
   \pnode(80,-20){backwards_in}
   \pnode(100,-10){backwards_out}

\psframe[framearc=.6](10,-35)(50,-50)
\rput(30,-42.5){$\Passive$}
\pnode(30,-35){passive_in1}
\pnode(30,-50){passive_in2}
\pnode(50,-42.5){passive_out}

   \psframe[framearc=1.0](60,-39)(100,-46)
   \rput(80,-42.5){$\Current$}
   \pnode(60,-42.5){current_in1}
   \pnode(80,-46){current_in2}
   \pnode(100,-42.5){current_out1}
   \pnode(80,-39){current_out2}
   \pnode(90,-46){current_out3}

        \psframe[framearc=.6](110,-35)(150,-50)
        \rput(130,-42.5){$\Active$}
        \pnode(130,-35){active_in1}
        \pnode(110,-42.5){active_in2}
        \pnode(110,-47.5){active_out1}
        \pnode(130,-50){active_out2}

\psframe[framearc=1.0](10,-69)(50,-76)
\rput(30,-72.5){$\New$}
\pnode(30,-76){new_in1}
\pnode(50,-72.5){new_in2}
\pnode(30,-69){new_out}

   \psframe(60,-65)(100,-80)
   \rput(80,-72.5){$\mathit{forward~simpl.}$}
   \pnode(90,-65){forwards_in}
   \pnode(100,-72.5){forwards_out1}
   \pnode(80,-65){forwards_out2}
   \pnode(60,-72.5){forwards_out3}

        \psframe(110,-65)(150,-80)
        \rput(130,-72.5){$\mathit{deduction~inf.}$}
        \pnode(130,-65){infer_in1}
        \pnode(120,-65){infer_in2}
        \pnode(130,-80){infer_out}

\pnode(30,-90){aux_node1}
\pnode(130,-90){aux_node2}

\ncline{->}{input_out}{passive_in1}\ncput{~~~1}
\ncline{->}{passive_out}{current_in1}\ncput{\parbox[c][20pt][t]{5pt}{2}}
\ncline{->}{current_out1}{active_in2}\ncput{\parbox[c][20pt][t]{5pt}{5}}
\ncline{->}{new_out}{passive_in2}\ncput{~~~8}
\ncline{-}{infer_out}{aux_node2}\ncput{~~~6}
\ncline{-}{aux_node2}{aux_node1}
\ncline{->}{aux_node1}{new_in1}\ncput{~~~6}

\ncline[linestyle=dashed]{->}{current_out2}{backwards_in}\ncput{~~~4}
\ncline[linestyle=dashed]{->}{active_out2}{infer_in1}\ncput{~~~6}
\ncline[linestyle=dashed]{->}{current_out3}{infer_in2}\ncput{~~~~~~~~~~6}
\ncline[linestyle=dashed]{->}{active_out1}{forwards_in}\ncput{\parbox[c][5pt][t]{40pt}{3,7}}

\ncline[doubleline=true]{->}{backwards_out}{active_in1}\ncput{~~~~~~~~4}
\ncline[doubleline=true]{->}{forwards_out2}{current_in2}\ncput{~~~~3}
\ncline[doubleline=true]{->}{forwards_out3}{new_in2}\ncput{\parbox[c][22pt][t]{5pt}{7}}

\end{pspicture}
                   \caption{\DISCOUNT\ algorithm
                         \label{fig:DISCOUNTAlgorithm}}

\end{center}
\end{figure*}

In the \DISCOUNT\ algorithm (see Figure~\ref{fig:DISCOUNTAlgorithm}), only 
active clauses can be 
simplified/discarded, or used to simplify/discard new clauses. 
So, there are no dashed lines between the box $passive$ and the forward and backward
simplification boxes. Note also that the clause in $current$ is subject to 
forward simplification (arrows labeled with~3), and it is used to simplify 
the active clauses (arrows labeled with~4). This is done to keep the set of active
clauses as simple \footnote{
Simplicity here is, of course, relative to the features of the used inference
system, in particular, the redundancy criteria.
} as possible.

In the \DISCOUNT\ algorithm passive clauses are constructed 
practically exclusively for evaluation of their properties which have to be known
for controlling the inference selection.
One may argue that the set of passive clauses is just a representation of all 
(potentially non-redundant) one-step inferences from the active clauses,
and from this point of view the \DISCOUNT\ algorithm implements
the idealistic notion of inference selection described in the beginning 
of Section~\ref{subsec:inference_selection_via_clause_sel}. 
In other words, the \DISCOUNT\ algorithm \emph{allows} 
the prover to observe
the space of all possible one-step inferences between active clauses,
which is a good thing by itself. 
However, the algorithm also \underline{\emph{obliges}} the system to do so 
by \emph{explicitly making all such inferences} and \emph{storing} 
the resulting clauses as passive. 
\emph{The cost of good inference selection becomes very high.} 
Typically, a thousand active clauses may generate
hundreds of thousands inferences, and a great deal of the resulting clauses
may be non-redundant with respect to the active ones, and, as such,
have to be stored as passive. Since the passive clauses are not
used for anything but selection, 
the work spent on constructing a clause may be \emph{frozen} 
for a long time, while the clause remains passive, and this work
is lost if the prover exhausts a given time or memory limit and
terminates. Storing huge numbers of passive clauses may additionally
require a lot of memory.

The \OTTER\ algorithm is not completely immune from any of these problems too.
In addition, the cost of simplification operations grows with the growth of
the set of passive clauses.

Recently there have been (at least) two attempts  to address some of 
these issues.
\Vampire\ implements the Limited Resource Strategy \cite{RiazanovVoronkov:JSC:LRS:2002}, which 
is intended to minimise the amount of work on generating, 
processing and keeping passive clauses in the \OTTER\ algorithm, 
which is wasted when the time limit is reached. This is done by discarding 
some non-redundant but heuristically bad clauses and inferences. 
\Waldmeister\ implements a sophisticated scheme to reduce the
memory requirements by the \DISCOUNT\ algorithm \cite{HL02,GHL03}.
In both cases, the adjustments of the top level algorithms
led to a great improvement in the effectiveness of the systems. 
This gives me hope that a radically different approach to inference selection 
may result in a real performance breakthrough.

\subsection{Finer selection units with graded activeness}

\myparagraph{Finer selection units.}
The inherent problems with the given-clause algorithm 
motivated me to look for a scheme that can 
\emph{facilitate better control of search} at an \emph{affordable cost}. 
Instead of selecting clauses, we are going to
\emph{select some particular parts (literals or subterms) 
of clauses 
and make them available for some particular kinds of inferences}. 
Such triples (clause + clause part + inference rule) will be the 
new selection units. This will help us to 
\emph{avoid premature invocation} of less promising clause parts. 

Stronger heuristics become available
for evaluating the quality of selection units since
such evaluation can take into account more than just integral
characteristics of a whole clause. For example, 
a selection unit with a generally good clause, but with a bad 
literal or subterm intended for a prolific
inference rule may now be given a low priority. 
On the one hand, this allows us to \emph{delay inferences with a bad part
of the clause}. On the other hand, \emph{we don't have to delay 
all inferences with the clause} simply because one of its parts
is bad. 

As an illustration, 
consider the unit clause $p(f(x,y),f(a,b))$. 
If some form of paramodulation is allowed, 
the subterm $f(x,y)$ is available for paramodulation into. 
This selection unit is \emph{extremely prolific}
since $f(x,y)$ unifies with all terms starting with $f$, 
and it makes a good sense to delay paramodulations into this
term without postponing other inferences with the clause,
e.g., paramodulations into $f(a,b)$.

Another example of a highly promising heuristic which 
is enabled by the proposed approach, is to give higher priority 
to binary resolution than to paramodulation, 
since the latter is often much more prolific than the former.
This heuristic already works very well (at least) in \Vampire.
The prover never enables inferences with positive equalities in a clause 
if there are literals of other kinds. Although generally successful,
this strategy often fails if all the other literals are relatively bad,
e.g., if they can generate many inferences. Consider
the clause $p(x,y) \vee q(a,b) \vee f(a,b) \simeq a$. The literal  
$p(x,y)$ is likely to be more prolific than $f(a,b) \simeq a$,
since $p(x,y)$ unifies with any atom starting with $p$. 
The proposed scheme allows us to give very high priority to $q(a,b)$,
lower priority to $f(a,b) \simeq a$ since this is a positive equality,
and a very low priority to the overly prolific literal $p(x,y)$.

Also, simplification inferences and redundancy tests 
can be treated in the same way as deduction inferences.
In the example above, we could make the term $f(a,b)$
available for rewriting immediately, and postpone
the integration of $f(x,y)$ into the corresponding
indexes until much later. By delaying simplification
inferences on stored clauses in a controlled manner we can 
achieve behaviours combining the properties of 
the \OTTER\ and \DISCOUNT\ algorithms. If simplification inferences
are given higher priority, the behaviour of our procedure
will be closer to that of the \OTTER\ algorithm. If simplification
inferences have priority comparable to the priority of deduction 
inferences such as resolution and paramodulation, we can expect
the new procedure to behave similar to the \DISCOUNT\ algorithm.

\myparagraph{Graded activeness.}
Apart from changing the subject of selection, \emph{I propose to
change the notion of selection itself}. The given-clause algorithm 
divides the search state in 
two parts. One part contains \emph{active} clauses, and the other one
contains \emph{passive} clauses
that are not yet available for deduction inferences. If  
a clause gets into the active set, it becomes available for all
future inferences regardless of its quality. 
To overcome this problem, I propose to use \emph{finer gradation
of selection unit activeness}. 

Intuitively, all selection units
would become \emph{potentially} available for inferences almost immediately, 
but some would be ``more available'' than the others. \emph{Less active 
selection units would be available for inferences with more 
active ones}.
High degree of activeness of a selection unit would indicate
higher priority of this unit for proof search. 
In the new procedure, the units containing parts of newly generated clauses 
initially receive the minimal degree of activeness, but later
are gradually promoted to higher degrees of activeness. 
When a promotion step takes place, the selection unit 
becomes available for new inferences with some units which
have not been eligible so far due to insufficient activeness. 
To give higher priority to inferences with heuristically better 
selection units,
the promotion frequency for different units should vary according
to their quality. Thus, we will be able to \emph{delay inferences between
heuristically bad inference units}. 

To illustrate this rather general scheme, I will outline a simple 
implementation scheme. For this implementation the nature of the used selection 
units is irrelevant, i.~e. they can be clauses as well as the finer selection
units proposed above. However, the implementation relies on the assumption 
that the quality of selected units is reflected by a special real-valued 
coefficient which takes positive values. 

If $\upsilon$ is a selection unit, 
the corresponding coefficient will be denoted as $quality(\upsilon)$. 
The intuitive meaning of the quality coefficient is the \emph{relative frequency
of promotion}. If $\upsilon_1$ and $\upsilon_2$ are two selection units,
at each promotion step the probability of selecting $\upsilon_1$
for promotion relates to the probability of selection $\upsilon_2$
as $quality(\upsilon_1)$ relates to $quality(\upsilon_1)$. 
Practically, we can select units for promotion randomly according to 
the distribution explicitly specified with the quality coefficients.
This selection discipline is known in the area of Genetic Algorithms as 
\emph{roulette-wheel selection} \cite{Goldberg:GeneticAlgorithms:1989}.

\begin{figure*}
\begin{center}
\psset{xunit=.135cm,yunit=.08cm,dash=5pt 4pt}
\begin{pspicture}(0,-20)(150,30)
\large
\psframe(0,0)(10,10)
\rput(5,4){$\Upsilon_0$}

\psframe(10,0)(20,10)
\rput(15,4){$\Upsilon_1$}

\psframe(20,0)(40,10)
\rput(30,4){$\ldots$}

\psframe(40,0)(50,10)
\rput(45,4){$\Upsilon_i$}

\psframe(50,0)(70,10)
\rput(60,4){$\ldots$}

\psframe(70,0)(80,10)
\rput(75,4){$\Upsilon_{n - i}$}

\psframe(80,0)(90,10)
\rput(85,4){$\Upsilon_{n - i + 1}$}

\psframe(90,0)(110,10)
\rput(100,4){$\ldots$}

\psframe(110,0)(120,10)
\rput(115,4){$\Upsilon_n$}

\normalsize

\pnode(5,15){pt1}
\pnode(35,15){pt2}
\ncline{->}{pt1}{pt2}
\rput(20,18){activeness increases}

\pnode(5,-2){pt3}
\pnode(5,-12){pt4}
\ncline{->}{pt4}{pt3}
\rput(8,-15){no inferences yet}

\pnode(45,-2){pt5}
\pnode(45,-12){pt6}
\pnode(115,-12){pt7}
\pnode(115,-2){pt8}
\ncline{-}{pt5}{pt6}
\ncline{-}{pt6}{pt7}
\ncline{->}{pt7}{pt8}

\pnode(85,-2){pt9}
\pnode(85,-12){pt10}
\ncline{->}{pt10}{pt9}

\pnode(100,-2){pt11}
\pnode(100,-12){pt12}
\ncline{->}{pt12}{pt11}

\large
\rput(92,-7){$\ldots$}
\rput(107,-7){$\ldots$}
\normalsize

\rput(80,-15){can have inferences with}

\pnode(45,22){pt13}
\pnode(115,22){pt14}
\ncline{-}{pt13}{pt14}

\pnode(45,12){pt15}
\ncline{->}{pt13}{pt15}

\pnode(115,12){pt16}
\ncline{->}{pt14}{pt16}

\pnode(60,22){pt17}
\pnode(60,12){pt18}
\ncline{->}{pt17}{pt18}

\pnode(75,22){pt19}
\pnode(75,12){pt20}
\ncline{->}{pt19}{pt20}

\pnode(85,22){pt21}
\pnode(85,12){pt22}
\ncline{-}{pt21}{pt22}

\pnode(100,22){pt23}
\pnode(100,12){pt24}
\ncline{->}{pt23}{pt24}

\rput(80,25){can have inferences with}

\end{pspicture}
               \caption{Graded activeness implementation
                 \label{sketch_for_graded_activeness_implementation}}
\end{center}
\end{figure*}
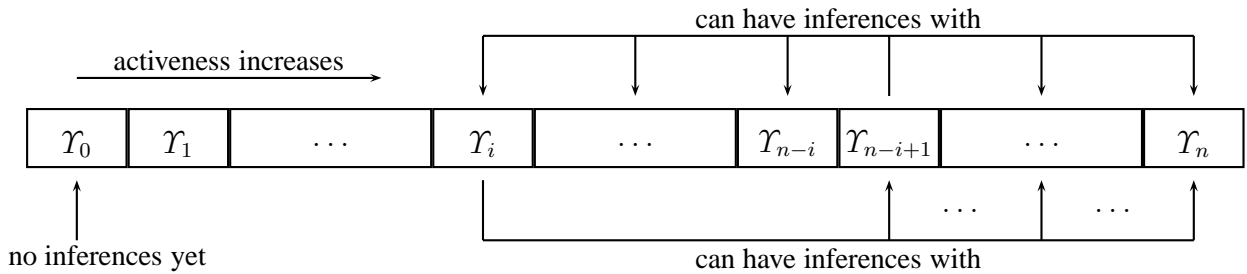

To realise the idea of graded activeness, I propose to partition 
the set of all available selection units into $n + 1$ sets 
$\Upsilon_0,\Upsilon_1,\ldots,\Upsilon_{n}$. The indexes of the sets
reflect the activeness of selection units contained in them: 
units in $\Upsilon_{i+1}$ are more active than units in $\Upsilon_i$.
More specifically, for $i > 0$, $\upsilon \in \Upsilon_i$ 
implies that \emph{all possible inferences between} $\upsilon$ 
\emph{and units from} $\Upsilon_{n-i+1},\ldots,\Upsilon_{n}$ 
\emph{have been made} and
\emph{no inferences between} $\upsilon$ \emph{and units in} 
$\Upsilon_0,\ldots,\Upsilon_{n-i}$ \emph{have been considered yet}.
$\Upsilon_0$ contains \emph{absolutely passive} selection units,
i.~e. units that \emph{have not participated in any inferences yet}.
This invariant, illustrated in 
Figure~\ref{sketch_for_graded_activeness_implementation}, 
is maintained by the following procedure. 
As soon as some selection unit is constructed, it is placed in $\Upsilon_0$.
At each macrostep of the procedure some selection unit $\upsilon$ 
is selected for promotion as outlined above. If $\upsilon$ happens to be 
in $\Upsilon_i$, where $i < n$, 
its \emph{promotion} means that $\upsilon$ is removed from
$\Upsilon_i$, all possible inferences between $\upsilon$ and selection units
from $\Upsilon_{n-i}$ are made, and $\upsilon$ is placed in $\Upsilon_{i+1}$.
Selection units from $\Upsilon_n$ are not promoted, they have the maximal
activeness.

Special arrangements may have to be made if we admit selection units 
that need not interact with other selection units to produce inferences.
For example, we may decide that selection units intended for binary factoring 
(\cite{BachmairGanzinger:HandbookAR:resolution:2001}) with a particular literal in a particular clause
do not need a counterpart unit, i.~e. if we decide to deploy such a unit, 
we will have to make all possible factoring inferences with the specified
literal within the specified clause. One possibility of dealing with 
such selection units is to designate some activeness $i > 0$ as 
a threshold, so that when a selection unit reaches $\Upsilon_i$,
all inferences requiring this unit alone are immediately made.

\vspace{5mm}
As a whole, the proposed inference selection scheme allows for much \emph{better control
over inference selection} which may translate into 
\emph{higher productivity of heuristics} and 
enables the use of \emph{new heuristics} 
which could not be used with the given-clause algorithm. 
Apart from other things, the extra flexibility\footnote{
The proposed design is strictly more flexible than the standard ones
since it is possible to implement it in such a way 
that both the \OTTER\ and \DISCOUNT\ algorithms can be simulated
by appropriate parameter settings.
} of inference selection will \emph{enhance the diversity of available strategies}\footnote{
My experience suggests that this is a very important factor
as in 2002--2005 the multitude of strategies supported by
the \Vampire\ kernel has been a major, if not the main, contributor to the 
growth of performance of the whole system.
}.
These advantages come at an \emph{affordable cost}.
The only involved overhead, caused by the need to store
large numbers of selection units, is compensated by lower
numbers of heuristically bad clauses which have to be created and stored
only to maintain completeness.

I would like to add one final consideration here. 
The calculi used in the state-of-the-art saturation-based provers
are designed with the aim of reducing search space. Partially, they
do this by restricting the applicability of resolution and
paramodulation rules. Often this is done by prohibiting inferences
with certain parts of clauses. For example, ordered resolution
with literal selection 
(see \cite{BachmairGanzinger:HandbookAR:resolution:2001}) 
prohibits resolving non-maximal positive
literals. However, restricting the shape of eligible derivations
also means restricting the number of eligible solutions, 
and \emph{simple solutions are often thrown away} if they do not satisfy 
the restrictions. 

It is possible, in principle, to relax
the restrictions by allowing some redundant inferences with 
some heuristically good parts of clauses.
For example, we may want to resolve large (i.~e., containing many symbols) 
positive non-maximal literals with the aim of obtaining smaller resolvents.
However, adjusting prover architectures based on the standard variants
of the given-clause algorithm makes very desirable the introduction of new 
ad hoc mechanisms for regulating the \emph{proportion of redundant
and non-redundant inferences}. 
The scheme proposed in this paper seems to have sufficient flexibility 
to accommodate such control mechanisms for free. For example, we can
allow selection units with large positive non-maximal literals,
and assign to them higher quality measure than 
to non-redundant selection units, if we are eager to derive small clauses
earlier. If we choose to be more conservative and want to avoid most
redundant inferences except a small number of very heuristically 
promising ones, we can always assign higher quality measure to
non-redundant selection units.

\subsection{Methodological considerations}

The proposed scheme for finer inference selection is completely compatible with the modern theory of resolution and paramodulation,
and requires no theoretical analysis. 
The difficult part of the job is to find an adequate design and to do the 
actual implementation. 

One of the implementation options is to adjust
an existing system. To investigate this possibility, I looked through 
the code of the kernel of \Vampire, v7.0, with the purpose of estimating the
amount of work required to adjust it to the new scheme. This investigation 
has convinced me that at least one third of the code would have to be
rewritten completely, and at least another third of it would have to 
be heavily adjusted to accomodate the new code. This is hardly surprising,
taking into account that the proposed changes target the top level design
as well as some key data representations and some mid-level functionality
such as indexing. 

The main conclusion of my inspection of the \Vampire\ kernel
code is that the amount of work required for a transition to the new scheme
is likely to exceed the cost of creating a rather advanced brand new 
prototype. An implementation from scratch can also be better tailored
to the new design. Considering this additional advantage,
my preference is clear. However, I do not dismiss the possibility
of implementing the new scheme on the base of other advanced saturation-based
provers.       

The nature of the proposed architectural principles is such that
their advantages can only be fully demonstrated if a significant 
effort is invested in design and assessment of search heuristics.
Indeed, the main advantages of the new inference selection approach
are the higher productivity of existing heuristics
and the possibility of using new heuristics. 

This extra flexibility in directing proof search can only be fully
exploited by means of \emph{tuning}.
Therefore, very extensive experimentation will be necessary
to find generally good combinations of parameters of heuristics,
as well as strategies specialised for important classes of problems
\footnote{
For experiments one can use the TPTP library
\cite{SutcliffeSuttner:TPTP2.4.1:2001}, which is at the moment 
the largest and most diverse collection of first-order proof problems.
It would also be very useful to look at more specialised large problem 
sets coming from applications in order to demonstrate the tunability
of the proposed architecture. 
}.
Strong tuning infrastructure to support such experimentation 
seems highly desirable. Developing such an infrastructure may be itself
an interesting research problem.

Finally, I would like to add a note about term indexing.
The finer gradation of activeness divides clauses and their parts
into many logically separate sets. An initial implementation may
adapt existing techniques to index these sets separately. 
However, better specialised indexing solutions may exist, and, if the proposed 
design proves viable in the initial experiments, it may
give rise to a new line of research in term indexing.


                 \section{Generalisation-based prioritising of search directions}


\subsection{Background: local syntactic relevancy estimation}

\emph{Blind} search in indefinitely large spaces is usually not effective
enough for most applications, so, all modern saturation-based provers 
try to predict the relevancy of particular search directions
by using various heuristics. In a saturation process state, 
the available search directions are identified by the accumulated clauses
(e.~g., the contents of the sets $\Passive$ and $\Active$ in the pick-given 
clause algorithm presented in Figure~\ref{fig:GivenClauseAlgorithm}).
The most common heuristics prioritise search directions by
giving some clauses higher priority for participating in inferences,
than the others. The estimation of relevancy of a clause is
based on such characteristics of the clause as its structural
complexity (e.g., simpler clauses get higher priority) or
its potential for participating in inferences 
(e.g., very prolific clauses get very low priority). 

Such approaches have natural limitations. The \emph{syntactic characteristics
of a clause}, used in the estimation, often \emph{fail to reflect 
the usefulness of the clause adequately}. 
For example, a structurally complex clause
may be absolutely indispensable for any solution of the problem at
hand, but it will be suspended for a long time.  
Another problem is that the estimation is done \emph{locally},
i.e. only one clause is analysed and \emph{global} properties of 
the current search state are not taken into account. 
For example, an absolutely irrelevant clause, i.e., 
participating in no minimal unsatisfiable subset of the current
clause set, may be given high priority because of its simplicity.

\subsection{Generalisation-based prioritising of proof-search directions}

To address the issues raised above, I propose a method for intelligent 
prioritising of search directions. 
The idea is as follows. We will \emph{estimate
the potential of a clause to participate in solutions of
the whole problem at hand by interacting with other 
currently available clauses}. 
Precise estimation is impossible
since it would require finding all, or at least some, solutions of the problem,
so we are looking for a good approximation. 

\myparagraph{General method.}
I suggest to 
\emph{probe the search space by exploring a substantially 
simpler search space}. 
The latter is obtained from the former by \emph{generalising
some search directions}. This is done by 
replacing (preferably large) clusters of similar clauses 
with their common generalisations. If we find a solution
of the simplified problem, which involves the generalisation
of a particular cluster, this is a good indication that at least some
of the clauses in the cluster \emph{can be} relevant. 
More importantly, \emph{the clauses whose generalisations 
have not yet proved useful, can be suspended as potentially irrelevant}. 
Additionally, the closer a resolved generalisation is to a particular clause
in its cluster, the better chances the clause has to participate
in a solution and the bigger priority it should be given.

Generalisations can be defined semantically:
a clause $C$ can be called a generalisation of clause $D$ if 
$C$ logically implies $D$.
For our purposes, however, it is convenient to use a simpler, syntactically
defined notion of generalisation, based on subsumption. In what follows,
we will call $C$ a generalisation of $D$ if $C$ subsumes $D$, 
i.~e. $C\theta \subseteq D$\footnote{More restrictive multiset-based 
variant of subsumption, where $C\theta$ is required to be a submultiset
of $D$, can also be used.}, where $C\theta$ and $D$ are viewed as sets
of literals.

\myparagraph{Implementation with naming and folding.}
Technically, the general approach described above can be realised by means
of a combination of \emph{dynamic naming} and \emph{folding}. This combination is 
called \emph{decomposition rule} in 
\cite{Hustadt+:DecompositionRule:LPAR:2005}, but for the purposes of this
paper it is convenient to consider the rules separately.
 
The idea should be clear from the following example. Suppose we have a clause
$C_1 = p(f(a,b)) ~\vee~ p(g(b,a)) ~\vee~ q(a)$.
We decide that this clause is too specific and its generalisation
$\Gamma_1(x_1,x_2) = p(f(x_1,x_2)) ~\vee~ p(g(x_2,x_1))$ should be
explored first. To this end, we introduce a new binary (according
to the number of variables in $\Gamma_1$)
predicate $\gamma_1$ and make it the \emph{name} for $\Gamma_1$. Logically, 
this can be viewed as introduction of the definition
$\forall x_1,x_2.~\gamma_1(x_1,x_2) \Leftrightarrow \Gamma_1(x_1,x_2)$.
We immediately transform $C_1$ by \emph{folding} this definition into the
following clause $C_1' = \gamma(a,b) ~\vee~ q(a)$.
Moreover, if there are other clauses, currently stored or
derived in the future, which are instances of the generalisation 
$\Gamma_1$, we can apply \emph{folding} to them as well, thus recognising
that the clauses are covered by the generalisation $\Gamma_1$. 
For example, if the clause 
$C_2 = p(f(h(a),b)) ~\vee~ p(g(b,h(a))) ~\vee~ r(b)$ 
is derived, it will be replaced by the clause $C_2' = \gamma_1(h(a),b) ~\vee~ r(b)$.
The generalisation $\Gamma_1$ is injected
into the search space in the form of the clause
$\Gamma_1(x_1,x_2) ~\vee~ \neg \gamma_1(x_1,x_2)$, which is a logical consequence
of the definition for $\gamma_1$.

In order to obtain the behaviour prescribed by the general scheme,
clauses containing \emph{$\gamma$-predicates}
(i.e., predicates which are generalisation names) are 
given special treatment. Namely, if a clause contains 
negatively the name $\gamma_1$ for the generalisation $\Gamma_1$, 
it means that the clause was derived from the clause 
$\Gamma_1(x_1,x_2) ~\vee~ \neg \gamma_1(x_1,x_2)$ representing the generalisation
$\Gamma_1$ in the search space. In such clauses, we prohibit all inferences
involving negative \emph{$\gamma$-literals} (i.e., literals with $\gamma$-predicates)
if there is at least one literal of a different kind.  
Roughly, in the clause $\Gamma_1(x_1,x_2) ~\vee~ \neg \gamma_1(x_1,x_2)$ 
we want to resolve the generalisation part 
$\Gamma_1(x_1,x_2)$ before we touch the literal $\neg \gamma_1(x_1,x_2)$.
Until this happens, the literal $\neg \gamma_1(x_1,x_2)$ only accumulates
the substitution which solves $\Gamma_1(x_1,x_2)$. 

When a clause containing only negative
literals with generalisation names is derived, this indicates that 
some generalisations ``fired'', i.e. they contradict each other 
and some ordinary input clauses. The derivation of such a clause 
can be viewed as a representation of a refutation of the clause
set consisting of the involved generalisations and
input clauses. We will call such clauses \emph{$\gamma$-contradictions}
and their inferences \emph{$\gamma$-refutations}.

Clauses containing positive $\gamma$-literals are \emph{suspended}
(temporarily removed from the search state)
\emph{until all of the corresponding generalisations have proved useful},
i.e. every participating $\gamma$-predicate belongs to at least
one $\gamma$-contradiction. When we can no longer suspend such a clause, 
we still block any inferences involving its non-$\gamma$ literals. 
A resolution inference between such a clause and a $\gamma$-contradiction
indicates that the clause is compatible with the corresponding
$\gamma$-refutation, and it represents an attempt to (gradually) \emph{refine}
the $\gamma$-refutation into a solution for the original problem. 
If some form of paramodulation is used, we have to allow
paramodulation into the positive $\gamma$-literals in an attempt
to make them compatible with available $\gamma$-contradictions.
 
To illustrate this, I continue the example. Suppose we have
derived the $\gamma$-contradiction $\neg \gamma_1(a,b)$. 
The clauses $C_1' = \gamma(a,b) ~\vee~ q(a)$ and 
$C_2' = \gamma_1(h(a),b) ~\vee~ r(b)$ can no longer be suspended. 
The clause $C_1'$ is directly compatible with the $\gamma$-refutation, 
which results in a derivation of the clause $q(a)$. 
The clause $C_2'$ is not compatible with the $\gamma$-refutation
since $\gamma_1(h(a),b)$ is not unifiable with $\gamma_1(a,b)$.
However, in presence of the unit equality clause $h(a) \simeq a$,
we can rewrite $C_2'$ into $\gamma_1(a,b) ~\vee~ r(b)$,
which is compatible with the $\gamma$-refutation, 
and then derive $r(b)$. Note, that the work spent on refuting the
generalisation $\Gamma_1$ (modulo some ordinary input clauses)
is utilised: we do not repeat the same inferences with the 
generalised literals from the original clause $C_1$. Moreover, the
results of this work are shared with another clause -- $C_2$, and,
potentially, with many other clauses covered by the generalisation 
$\Gamma_1$. Such sharing of work on similar parts of potentially very many 
different clauses can be an additional advantage. 

Note that the proposed naming- and folding-based scheme is 
rather flexible. It allows many variants which may differ, e.g., in
the way suspended clauses are treated, how selection of inferences
is done with the $\gamma$-literals, how generalisations are chosen, 
how many generalisations can be applied to a single clause and
whether they can be overlapping \footnote{Intuitively, two generalisations 
of a clause $C$ overlap if they cover some common literals in $C$. For example,
$C_1 = p(f(a,b)) ~\vee~ p(g(b,a)) ~\vee~ q(a)$ has overlapping
generalisations $p(f(x_1,x_2)) ~\vee~ p(g(x_2,x_1))$ 
and $p(g(x_1,x_2)) ~\vee~ q(x_2)$ because the literals $p(g(x_2,x_1))$
and $p(g(x_2,x_1))$ both generalise the literal $p(g(b,a))$.
}, etc. The description above is only intended to provide a general framework
for formulating such variants. Moreover, it is obviously not the only
possible framework for implementing the general scheme presented
in the beginning of this section.   

The proposed implementation scheme offers another advantage for
free. The user gets an additional means of controlling proof search
by specifying in the input which clauses he would like to make 
named generalisations from the start. This can be viewed as
a way of hinting at useful lemmas (of a restricted kind since only
clauses are named rather than arbitrary formulas) or suppressing 
search directions which do not seem promising to the user. 

For example, by analysing some previous proof attempts the user
may conclude that many clauses of the form 
$\neg p(g(a,b)) ~\vee~ C$ are generated. 
If the user has reasons to believe that the literal 
$\neg p(g(a,b))$ can be solved, i.~e.  $p(g(a,b))$ is logically
implied by the input clauses, he may want to try proving $p(g(a,b))$
as a lemma, and later use it to resolve with the clauses 
$\neg p(g(a,b)) ~\vee~ C$. Practically, this can be done by making
$\neg p(g(a,b))$ a generalisation and giving it some name, e.~g. 
$\gamma_3$. Refuting $\neg p(g(a,b)) ~\vee~ \neg \gamma_3$ corresponds
to proving the lemma $p(g(a,b))$, and resolutions between 
the $\gamma$-contradiction $\neg \gamma_3$ and clauses of the form
$\gamma_3 ~\vee~ C$ correspond to applications of the lemma.
Such lemma hinting may be beneficial because it allows to share
the work on solving literals $\neg p(g(a,b))$ in many different clauses
instead of solving them separately.

If the user has reasons to believe that the literal $\neg p(g(a,b))$
cannot be solved, and thus all the clauses $\neg p(g(a,b)) ~\vee~ C$
are redundant, it still makes sense to make $\neg p(g(a,b))$ a generalisation.
This will keep the generalised clauses $\neg p(g(a,b)) ~\vee~ C$ away
from inferences without completely discarding them. Only if the user's
intuition was incorrect, i.~e. $\neg p(g(a,b))$ can actually be solved,
the generalised clauses are reintroduced in the search space.

\subsection{Related work}

\myparagraph{Static relevancy prediction.}
My original idea was to use some sort of clause abstractions
for dynamic suppressing of potentially irrelevant search directions
in the framework of saturation-based reasoning. This idea
was inspired by \cite{FuchsFuchs:CADE:relevancy:1999} where the authors
propose to use various clause abstractions for statically identifying 
input clauses which are \emph{practically irrelevant}, i.e. can not
be useful in a proof attempt of acceptable complexity. 
Roughly, this is done by applying abstractions to an input clause set, 
exploring the space of all proofs of restricted complexity
with the abstracted clause set, and throwing away the input clauses
whose abstractions do not participate in any of the obtained 
proofs with the abstracted set.

\myparagraph{Iterative generalisation-refinement.}
Some time ago \cite{Plaisted:CADE:generalisations:1986} drew my 
attention to the simplest kind of clause abstractions 
-- generalisations, which seems convenient for our purposes.
The method works roughly as follows. A resolution prover is
parameterised by a generalisation function on clauses, i.e. a function
which computes several, possibly overlapping, 
generalisations for a given clause. 
When the prover is run on a problem, the generalisation mechanism
replaces suitable clauses by their generalisations. 
The whole scheme works as iteration through levels of generalisation
strength. First, the prover is run with a strong generalisation
function to enumerate all refutations with depth below a certain
limit. Then the generalisation function is weakened\footnote{
Roughly, a weaker generalisation function produces more specific
generalisations of a given clause.
} and the prover 
uses the previously found refutations to guide the enumeration 
of refutations with the new generalisation function. The key idea
is that the refutations with the weaker generalisation function
are in a certain (strict) sense refinements of the refutations
obtained with the stronger generalisation function. 
Such refinement is performed repeatedly and at some point
the prover tries to refine a refutation from the previous step
into a refutation which uses no generalisation.

\myparagraph{Octopus approach.} The \Octopus\ system 
\cite{NewbornWang:JAR:Octopus:2005}
runs a large number of sessions of the prover 
\Theo\ \cite{Newborn:ATP:2001} distributed over a cluster of computers.
Each \Theo\ session first runs on a \emph{weakening} of the original problem,
obtained by 
\emph{replacing one of the clauses with one of its generalisations}.
If one of the sessions succeeds in solving the weakened problem,
the solution is used to direct the search for a solution of the original
problem in two ways:

\begin{itemize}

  \item The unmodified clauses from the original problem formulation,
 	which participate in the solution of the weakened problem, 
	are considered to be heuristically more relevant. In the future
	searches for solutions of the original problem, these clauses 
	are given higher priority.

  \item Some clauses in the obtained refutation of the generalised clause set,
        which were derived from unmodified clauses, are added as lemmas to the 
        problem formulation.
\end{itemize}

\vspace{5mm}
The main difference between my approach and the static relevancy prediction 
approach of \cite{FuchsFuchs:CADE:relevancy:1999}, and also 
the Octopus approach \cite{NewbornWang:JAR:Octopus:2005}, is that our clause 
generalisations are introduced \emph{dynamically}, and
\emph{can be used on derived clauses}. This allows a good degree of 
\emph{adaptivity}. 

My approach is closer to, and can be viewed as an attempt
to revive the line of work presented in \cite{Plaisted:CADE:generalisations:1986}.
I hope to improve on this approach mainly by enumerating generalised
refutations \emph{lazily}, thus \emph{avoiding any artificial 
limits on the complexity of refutations} and the need to
enumerate a whole, potentially large, set of generalised refutations
before we try to use these refutations.
Also, my approach is 
more \emph{semantic} in its nature since we do not try to refine 
generalised refutations by following their structure. We are
interested in \emph{existence} of $\gamma$-refutations rather than
their shape. This allows much easier integration with various
variants of resolution- and superposition-based inference systems.
Additionally, my approach imposes no restrictions on how the
generalisation functions are specified and implemented. In particular,
the generalisation mechanism can be adaptive. For example, 
the strength of generalisation may depend on various properties of
clauses being generalised, or even on some global properties of the current
search state. 

The general method is also partially inspired by, and 
shares some philosophical ideas with \cite{Plaisted:AI:Abstraction:1981} 
and \cite{Giunchiglia+:AICOM:abstractions:1997}.

The use of naming and folding is a natural continuation 
of our joint work with Andrei Voronkov on implementing \emph{splitting
without backtracking} \cite{RiazanovVoronkov:IJCAI:Splitting:2001} 
and also partially stems from 
an unfinished attempt by the author to mimic 
\emph{tableaux without backtracking} 
\cite{Giese:IJCAR:IncrementalClosure:2001}
in the context of saturation. 
Recently I have discovered that \cite{Hustadt+:DecompositionRule:LPAR:2005} 
proposes to use exactly the same combination of naming and folding, 
under the name of \emph{decomposition rule}, for deciding two description
logics and query answering in one of them.

{\bf Semantic guidance in the style of \SCOTT.} 
To conclude the overview of relevant work, I would like to mention another 
approach which is technically unrelated to the one proposed here, 
but which also provides an alternative
to local syntactic relevancy estimation. 

The \emph{semantic guidance}
approach, developed within the \SCOTT\ project 
\cite{HodgsonSlaney:AICOM:SCOTT:2002}, is roughly as follows.
The prover tries to establish satisfiability of several sets 
of stored clauses (in \SCOTT\ this is done with the help of an 
external model builder). Ideally, these sets must approximate their 
maximal satisfiable supersets as closely as possible. The sets are
used for guiding clause selection roughly as follows: 
clauses participating in fewer such satisfiable sets are given higher 
priority for selection. The intuition behind this approach is
that a clause is more likely to be redundant if it participates
in many satisfiable sets. This heuristic is supported by 
the fact that if a clause is in every \emph{maximal} consistent subset, 
then it is definitely redundant. 

The applicability of the semantic guidance approach seems limited because 
it relies on the costly operation of establishing satisfiability 
of large clause sets.
This overhead may be acceptable in solving very hard problems when 
the user can afford to run a prover for hours or even days. 
Many applications, however, require solving large numbers of simpler problems
and much quicker response. I hope that generalisation-based guidance
can be more useful for this kind of applications because the associated 
overhead seems more manageable due to the flexibility of
generalisation function choice. Anyway, a meaningful comparison of the
two approaches can only be done experimentally, when at least one variant
of the generalisation-based method is implemented.

\subsection{Methodological considerations}

Certain theoretical effort is required to formulate the method in 
full detail. It makes sense to consider a number of variants of the method and
try to predict their strengths and weaknesses.
It is also essential to have a clear picture of how the proposed use 
of generalisations will interact with the popular inference systems
based on resolution, paramodulation and standard simplification
techniques. In particular, it is necessary to consider the 
search completeness issues. 

The effectiveness of the method is likely to depend strongly
on the choice of generalisation functions and, therefore, 
a significant effort to find adequate heuristics would be well justified.
In particular, anybody implementing the method is very likely to 
encounter the problem of \emph{overgeneralisation}. Working with too strong
generalisations of clauses may potentially lead to numerous 
$\gamma$-refutations that are not compatible with any of the covered
clauses. For example, if we fold the definition   
$\forall x. \gamma(x) \Leftrightarrow p(x)$ into the clause 
$p(f(a))$, transforming it into $\gamma(f(a))$,
we may later derive some $\gamma$-refutation $\neg \gamma(b)$ which 
is incompatible with $\gamma(f(a))$. The work on deriving $\neg \gamma(b)$
is potentially wasted, unless, of course, there are other clauses
compatible with $\neg \gamma(b)$.
Another problem with overgeneralisation is that $\gamma$-refutations
compatible with many clauses may be found quickly, and will activate 
the corresponding clauses. In such cases
the work spent on creating generalisations themselves and their application 
to clauses, is wasted because the generalisations do not fulfill
their mission of suspending clauses. On the other hand, too weak 
generalisations may also be bad, e.~g., because they cover too small 
sets of clauses, in which case their construction is not properly amortised.
I hope these considerations illustrate the thesis about importance of 
searching for heuristics for choosing effective generalisation functions.

In contrast with the fine inference selection scheme 
which essentially requires creating a new implementation, 
the generalisation-based search guidance can be relatively
easily integrated into some existing provers, especially
if it is implemented with naming and folding as outlined earlier.
My experience with implementing splitting-without-backtracking
\cite{RiazanovVoronkov:IJCAI:Splitting:2001} (see also
Chapter 5 in \cite{Riazanov:PhDThesis:2003}) in the \Vampire\ kernel
suggests that only a moderate effort is required to implement
naming and folding on the base of a reasonably manageable implementation
of forward subsumption, which is a standard feature in advanced saturation-based provers. 

The most difficult task is likely to be the design and implementation 
of a flexible, yet manageable, mechanism for specifying generalisation 
functions, and to provide a higher-level interface for this mechanism
which would enable productive use of heuristics. 
The reliance on heuristics also implies that very extensive 
experimentation will be required to assess the general effectiveness 
of the method and to compare its variants.


                 \section{Acknowledgments}


This paper is almost entirely based on my work on \Vampire\ 
in the Computer Science Department at the University of Manchester.
The work was supported by a grant from EPSRC. 
The first draft of this paper was also written in Manchester.
I would like to thank Andrei Voronkov for useful discussions 
of the ideas presented here. Many thanks to Geoff Sutcliffe
for his scribblings on the first draft of this paper.


\bibliographystyle{plain}
\bibliography{new_implementation_framework}

\end{document}